\documentclass{article} 
\usepackage{colm2024_conference}

\usepackage{microtype}
\usepackage{hyperref}
\usepackage{url}
\usepackage{booktabs}
\definecolor{darkblue}{rgb}{0, 0, 0.5}
\hypersetup{colorlinks=true, citecolor=darkblue, linkcolor=darkblue, urlcolor=darkblue}
\usepackage{graphicx}
\usepackage{subfigure}
\usepackage{multirow}
\usepackage{xfrac}    
\usepackage[most]{tcolorbox}
\usepackage{color}
\definecolor{keywordcolor}{rgb}{0.7, 0.1, 0.1}   
\definecolor{tacticcolor}{rgb}{0.0, 0.1, 0.6}    
\definecolor{commentcolor}{rgb}{0.4, 0.4, 0.4}   
\definecolor{symbolcolor}{rgb}{0.0, 0.1, 0.6}    
\definecolor{sortcolor}{rgb}{0.1, 0.5, 0.1}      
\definecolor{attributecolor}{rgb}{0.7, 0.1, 0.1} 

\title{InternLM-Math: Open Math Large Language Models Toward Verifiable Reasoning}
\colmfinalcopy

\author{
    \textbf{Huaiyuan Ying\textsuperscript{1,2*}, Shuo Zhang\textsuperscript{1,3\thanks{Equal contribution.}}, Linyang Li\textsuperscript{3}, Zhejian Zhou\textsuperscript{1,4}, Yunfan Shao\textsuperscript{1,3}} \\
    \textbf{ Zhaoye Fei\textsuperscript{1,3}, Yichuan Ma\textsuperscript{1}, Jiawei Hong\textsuperscript{1,3}, Kuikun Liu\textsuperscript{1}, Ziyi Wang\textsuperscript{1}, Yudong Wang\textsuperscript{1}}\\ 
    \textbf{ Zijian Wu\textsuperscript{1,5}, Shuaibin Li\textsuperscript{1}, Fengzhe Zhou\textsuperscript{1}, Hongwei Liu\textsuperscript{1}, Songyang Zhang\textsuperscript{1}} \\ 
    \textbf{ Wenwei Zhang\textsuperscript{1}, Hang Yan\textsuperscript{1}, Xipeng Qiu\textsuperscript{3}, Jiayu Wang\textsuperscript{1}, Kai Chen\textsuperscript{1}, Dahua Lin\textsuperscript{1}} \\
    \\
    {\textsuperscript{1} Shanghai AI Laboratory} \\ 
    {\textsuperscript{2} Tsinghua University} \\
    {\textsuperscript{3} Fudan University, School of Computer Science} \\
    {\textsuperscript{4} University of Southern California} \\
    {\textsuperscript{5} Shanghai Jiaotong University} \\
    \\
    \texttt{internlm@pjlab.org.cn} \\
}

\begin{document}

\maketitle

\begin{abstract}
The math abilities of large language models can represent their abstract reasoning ability.
In this paper, we introduce and open-source our math reasoning LLMs InternLM-Math which is continue pre-trained from InternLM2.
We unify chain-of-thought reasoning, reward modeling, formal reasoning, data augmentation, and code interpreter in a unified seq2seq format and supervise our model to be a versatile math reasoner, verifier, prover, and augmenter. These abilities can be used to develop the next math LLMs or self-iteration.
InternLM-Math obtains open-sourced state-of-the-art performance under the setting of in-context learning, supervised fine-tuning, and code-assisted reasoning in various informal and formal benchmarks including GSM8K, MATH, Hungary math exam, MathBench-ZH, and MiniF2F.
Our pre-trained model achieves 30.3 on the MiniF2F test set without fine-tuning.
We further explore how to use LEAN to solve math problems and study its performance under the setting of multi-task learning which shows the possibility of using LEAN as a unified platform for solving and proving in math.
Our models, codes, and data are released at \url{https://github.com/InternLM/InternLM-Math}.
\end{abstract}
\def\lstlanguagefiles{lstLEAN.tex}

\begin{center}
Demo: \url{https://huggingface.co/spaces/internlm/internlm2-math-7b}
\end{center}

\section{Introduction}
Large language models \citep{gpt3,Minerva,GAL,gpt4,palm2,2023internlm,llemma,gemini,shao2024deepseekmath} have shown significant abilities in mathematical reasoning tasks from grade school \citep{gsm8k} to high school levels \citep{MATHbenchmark} by using chain-of-thought reasoning \citep{cot} or program-of-thought reasoning \citep{pot,pal}.

Building such models requires pre-training on math corpora and supervised fine-tuning on math problems.
We introduce InternLM-Math based on InternLM2-Base models\footnote{\url{https://github.com/InternLM/InternLM}}.
InternLM2 shows strong performance in various aspects including math, code, chat experience, instruction following, and creative writing.
We retrieve and collect math-related data to continue pre-training on InternLM2-Base and obtain state-of-the-art performance on informal and formal math reasoning benchmarks outperforming Minerva \citep{Minerva} and Llemma \citep{llemma}.

During supervised fine-tuning, we supervise InternLM-Math not only on solving math problems using chain-of-thought and code interpreters but also many tasks for developing math LLMs which include reward modeling and augment helper. We also introduce using LEAN for translating between natural languages and LEAN statements \citep{pact}, solving easy math problems, and proving math statements.
InternLM-Math series models achieve open-sourced state-of-the-art performance on multiple benchmarks \footnote{This work is concurrent with Deepseek-Math.} and score more than 90\% relative to GPT-4 \citep{gpt4}. 

Our contributions including:
\begin{itemize}
    \item We open-source our base and SFT LLMs in math reasoning. It achieves open-sourced SOTA under the setting of ICL, SFT, RM reranking, and Python-assisted in various benchmarks.
    \item We unify chain-of-thought reasoning, reward modeling, data augmentation, and formal reasoning under a unified seq2seq format. We supervise our model with both problem-solving and verification abilities.
    \item We propose reasoning interleaved with coding (RICO) and achieve state-of-the-art math reasoning with Python's assistance.
    \item We explore using LEAN to solve math word problems and we investigate its performance concerning data size during multi-task learning.
\end{itemize}

\begin{figure}[htbp]
 \centering
 \includegraphics[keepaspectratio, scale=0.2]
      {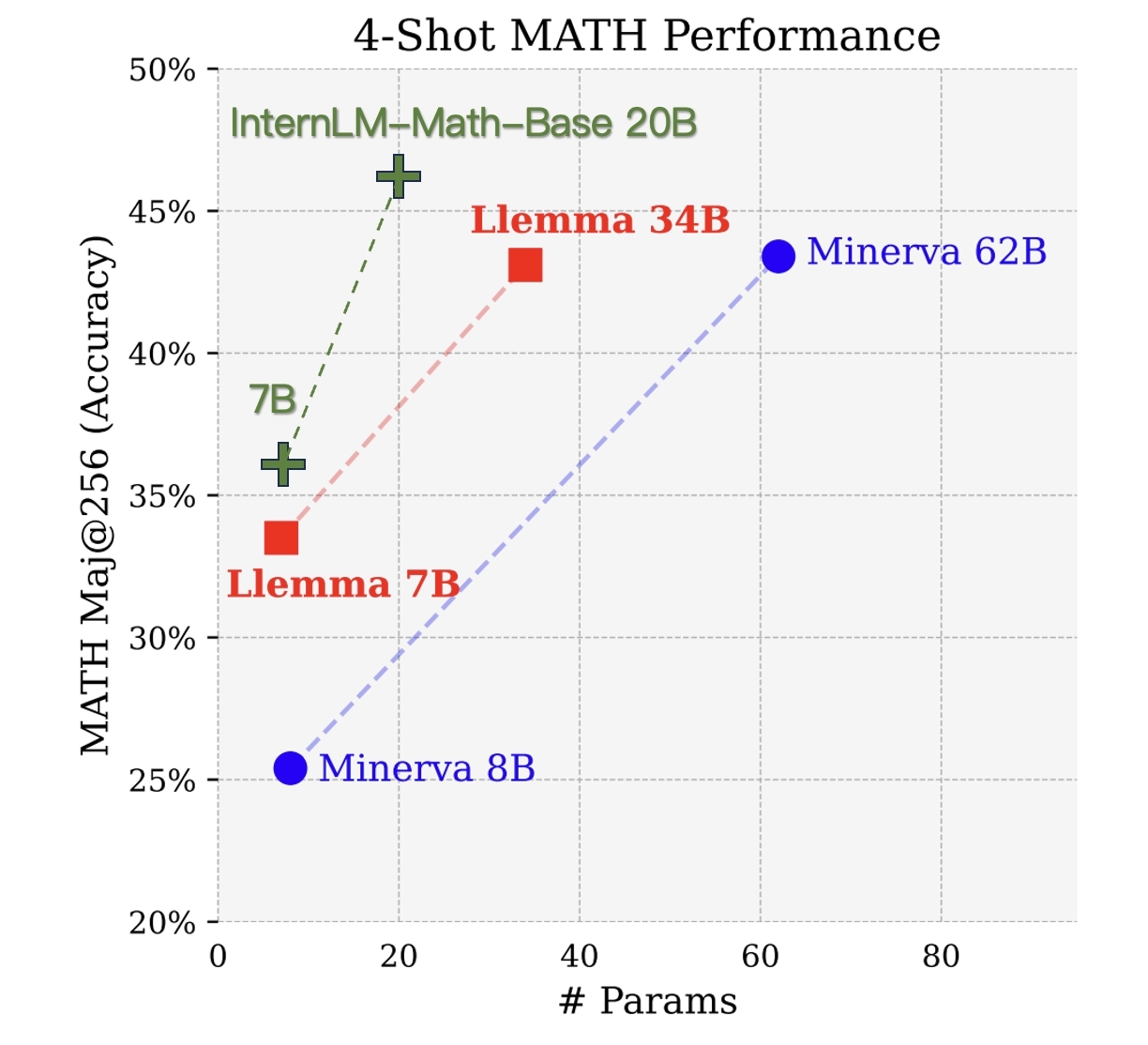}
 \caption{4-shot MATH performances with 256 times majority voting. Comparison is based on our pre-trained base model, Llemma \citep{llemma}, and Minerva \citep{Minerva}. The figure is modified from \cite{llemma}.}
 \label{fig:head}
\end{figure}

\section{Related Work}
\paragraph{Math Pre-training} Pre-training helps LLMs acquire computational and mathematical knowledge from various sources, such as math corpora \citep{pact,Minerva,openwebmath,wang2023generative}, problem sets \citep{lightman2023lets}, and synthetic data \citep{hendrycks2021measuring,goat,mathglm}.
ArXiv with abundant math contents is usually used in math pre-training \citep{lewkowycz2022solving,GAL,azerbayev2023proofnet}.
\cite{openwebmath} extracts math web pages from common crawl which can be complementary to arXiv.
Math problems including GSM8K \citep{gsm8k} and MATH \citep{MATHbenchmark} are used during pre-training \citep{gpt4}.
Synthetic problems generated via rules or math program scripts \citep{hendrycks2021measuring} can also be used for pre-training.
InternLM-Math collects pre-train data from math corpora and synthetic data which establish its math ability.

\paragraph{Math Fine-tuning} 
Building a stronger augmented chain-of-thought dataset \citep{yu2023metamath,yue2023mammoth,mugglemath,liu2024augmenting} for SFT to improve math reasoning performance has received lots of interest.
Problem augmentation \citep{luo2023wizardmath,yu2023metamath,mugglemath,liu2024augmenting} and reasoning path augmentation \citep{zelikman2022star,huang2022large,core,rft} are two common ways which show significant effectiveness.
Reinforcement learning \citep{uesato2022solving,luo2023wizardmath,shepherd,Singh2023BeyondHD} has also been explored to fine-tune language models to lift reasoning performance which requires a reward model (i.e. verifier) that can distinguish correct and incorrect reasoning processes and answers.
Compared to previous work, we not only build a stronger chain-of-thought SFT dataset but also integrate multiple abilities all-in-one including verification, proof, code interpreter, and data augment helper.

\paragraph{Math Verifier} 
Reward models (verifiers) are usually used for reranking multiple candidate reasoning paths.
Outcome reward models (ORM) \citep{gsm8k} and process reward models (PRM) \citep{uesato2022solving,lightman2023lets} leverage LLMs to verify the correctness of answers and processes.
ORM is less efficient than PRM \citep{lightman2023lets,shepherd} while PRM requires many human experts for labeling.
To reduce human labor in labeling PRM, \cite{ovm,shepherd} determine the correctness of processes based on multiple completions from the process.
Compared to previous work which trained two separate networks for the policy network and the reward model.
We combine these two functions into one unified seq2seq format. After SFT, our model can be used for solving problems or verifying reasoning paths.
Furthermore, we explore leveraging formal math language LEAN to verify reasoning paths by translating to LEAN codes.

\paragraph{Math Code Interpreter}
Code interpreters can complement LLMs with complex calculation capabilities via various Python libraries.
Early explorations use program-of-thought \citep{pot, pal} but lack interaction between LLMs and tools, which may not be able to handle scenarios that require multi-step reasoning and calculation. Recent works \citep{gou2023tora, wang2023mathcoder} try to more seamlessly integrate code interpreter with reasoning by allowing the model to summarize the results based on the one-time code execution outcomes.
InternLM-Math explores reasoning interleaved with coding (RICO), \textit{i.e.}, the reasoning and coding processes are interleaved for multiple rounds until the problem is solved, which is more natural and close to the problem-solving process and fully exploits the reasoning capabilities of LLM.

\paragraph{Math Prover} 
Solving a math problem with a correct answer by LLMs still cannot ensure its process's correctness. However, proving a statement using formal languages like Isabelle \citep{paulson2000isabelle}, LEAN \citep{lean}, and Coq \citep{Coq-refman} can promise it.
Training an LLM to automated theorem proving in formal languages \citep{pact,curricum,azerbayev2023proofnet,yang2023leandojo,llmstep,lyra} is hard due to sparse data.
InternLM-Math achieves state-of-the-art few-shot performances on MiniF2F \citep{zheng2021minif2f} which shows potential in building a strong math prover.

\section{Pre-training}

In this section, we first describe our pre-training data composition. Then, we outline our data post-processing method we perform on the training data. Finally, we dive into the details of our training strategy.

\subsection{Pretrain data composition}

To achieve a competitive performance in mathematical domains, we collected a diverse collection of high-quality data. 
We do not leverage any LLM-generated data during pre-training.
This data is systematically categorized into the following categories:

\paragraph{CC Retrieved Data} We employed Query of CC \citep{2024arXiv240114624F} to retrive the training corpus for InternLM2-Math-Base.
We select math-related corpus from Query of CC corpus as the first part of continue pretraining data. This part includes 20B tokens.

\paragraph{Domain-Specific Data} We selected from open-source dataset \citep{llemma} and in-house high-quality datasets in the field of mathematics including web pages, codes, arXiv, forums, and books. This part includes 11B tokens.

\paragraph{Synthetic Data} 
We synthesized numerical operation data to improve the model's numerical operation capabilities. 
We included five common operations including arithmetic, exponentiation, logarithmic, trigonometric, and polynomial calculations.
For numerical operations, we traversed a set of commonly used values and randomly sampled a wider range of values within 10-digit numbers.
To prevent the model from overfitting to specific templates, we diversely constructed templates to ensure the model's numerical computing capabilities generalize to some extent. This part includes 0.2B tokens.

\subsection{Data post-processing}

To enhance the quality of the training data, we follow the approach of Query of CC\citep{2024arXiv240114624F} and implement a series of data post-processing strategies. Specifically, we trained a scoring model to identify high-quality datasets. Subsequently, We used the Minhash-LSH method for deduplication of the training data. In our practice, we filtered out duplicate data with a similarity exceeding 0.7.

We further conducted exact formulation decontamination for the special domain data on the MATH test set. We extract all formulations within a given paragraph. If the concatenation of formulations hits any in the MATH test set, we simply remove them.

\begin{table}[]
\centering
\begin{tabular}{ccccc}

\toprule  
Domain&Dataset& Unique Tokens(B)& Epochs &Tokens(B)\\
\midrule  
\vspace{5px}
\multirow{1}{*}{CC Retrieved Data} &knowledge pile& 20& 4& 80\\

\multirow{3}{*}{Special Domain Data}&open-web-math& 6& 4& 24\\
&algebraic-stack& 4& 4& 16\\
\vspace{5px}
&others& 1& 4& 4\\
Synthetic Data&num& 0.2& 5& 1\\
\midrule  
Total& -& 31.2&-&125\\
\bottomrule 
\end{tabular}
\caption{Pre-train data usage for InternLM-Math. Unique Tokens refers to the number of tokens in the original dataset. Tokens refers to the total number of tokens in the dataset consumed during pre-training.}
\label{table_1}
\end{table}
\subsection{Training strategy}

After collecting and post-processing high-quality retrieved and domain-specific data, we set different training epochs. For details of the data ratios, see Table \ref{table_1}. In total, we collected 31.2 billion high-quality tokens across the datasets, we follow \cite{muennighoff2023scaling} to use 4 epochs for most datasets.
To continue pre-training on InternLM2-Base, we adopt the same training strategy across different model sizes, as described in InternLM \citep{2023internlm}, and we use InternEvo\footnote{https://github.com/InternLM/InternEvo} as the training framework. During training, we used a context length of 4096. For documents that are too long or too short, we either truncate or concatenate them to achieve the desired context length. We adopt mixed-precision training with bfloat16 and FlashAttention2 \citep{dao2023flashattention2} to attain optimal memory utilization and training efficiency. The standard AdamW \citep{2017arXiv171105101L} optimizer was employed with hyperparameters $\beta_1=0.9$, $\beta_2=0.95$, and $weight\_decay=0.1$. We use a standard cosine learning rate scheduler. Specifically, the model's learning rate reaches a maximum of $lr_{max}=3e-5$ after 2000 warm-up steps, and then gradually decreases to a minimum of $lr_{min}=3e-6$ over the course of training. In the continued pre-training process, a total of 125 billion tokens were trained.
For the 20B model, we early stop at 80 billion tokens based on in-context learning performance.

\begin{figure}[t]
 \centering
 \includegraphics[keepaspectratio, scale=0.45]
      {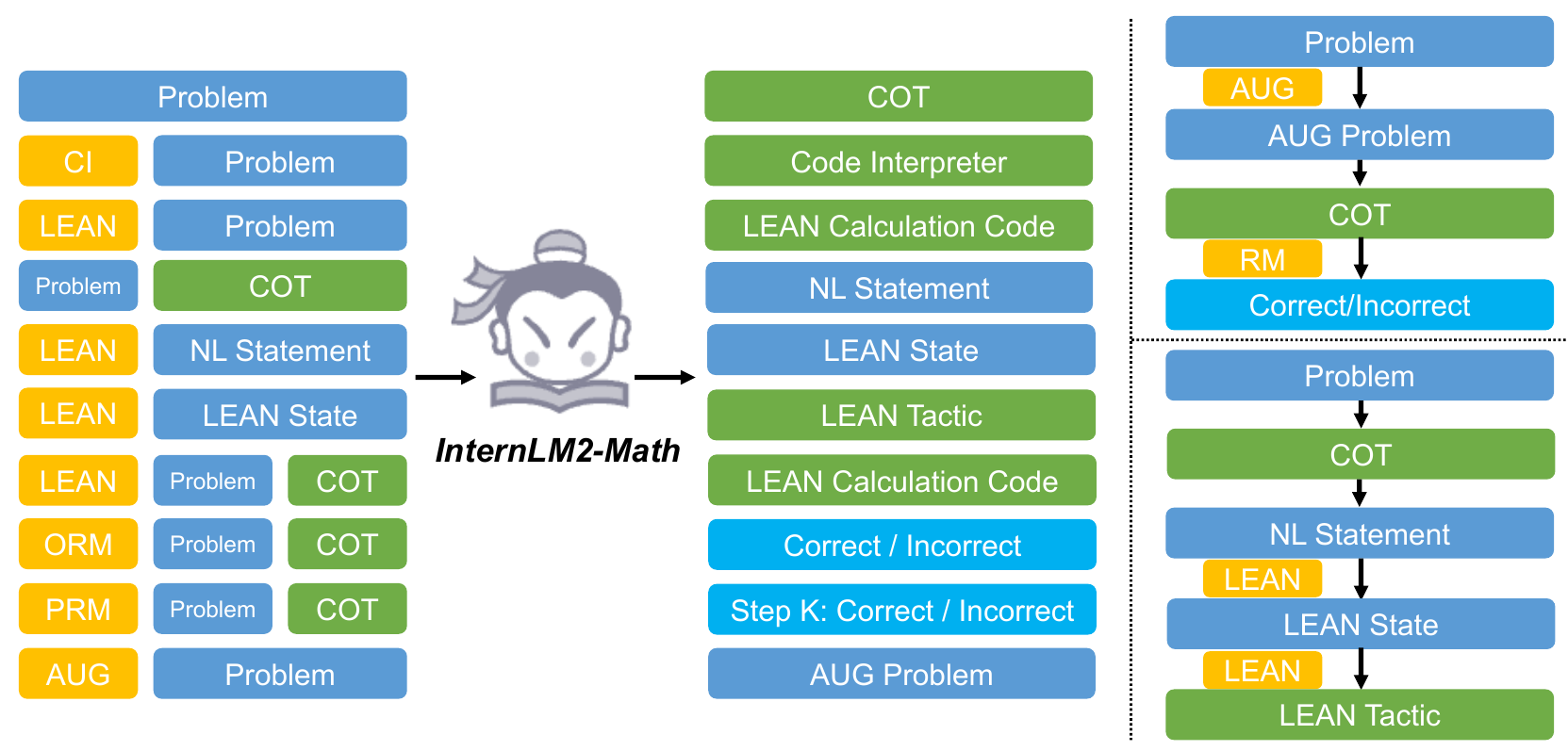}
 \caption{The left part shows the query and response formats in SFT. The right parts show two possible usages of our SFT model. The upper right is a pipeline of synthesizing new problems using our augment helper, COT, and RM abilities. The lower right is a pipeline of solving informal problems using formal languages with COT and formal reasoning abilities.}
 \label{fig:head_sft}
\end{figure}

\section{Supervised Fine-tuning}
Dislike other math-specialized LLMs that focus on solving math problems, our models are targeted to be math solvers and also be ready for self-improving which requires abilities including problem augmentation, reward modeling, self-verifying, formal reasoning, and code interpreters. 
Our SFT data contains high-quality human-written, rule-generated, and LLM-generated data for the abovementioned abilities,
the detailed data composition can be seen in Table~\ref{tab:SFT-data-composition} and our models generated responses can be seen in Appendix~\ref{appendix:sft_casestudy}.
We show the query and response format of SFT in Figure~\ref{fig:head_sft}.

\paragraph{Chain-of-thought}

We utilize MetaMath \citep{yu2023metamath} as our fundamental English chain-of-thought data resource which brings consistent reasoning improvement to various LLMs.
We leverage in-house Chinese datasets for Chinese chain-of-thought abilities.
To improve the math reasoning abilities of our models' weaknesses, we apply reasoning path augmentation \citep{zelikman2022star} on specific datasets.
Inspired by \cite{uesato2022solving}, we use final answers and our trained PRM to filter augmented reasoning paths.

Although we add synthesis calculation data in pre-training, the models still perform badly on complex calculations.
In the spirit of scratchpad \citep{scrachpad,goat}, we decompose complex calculations including multiple brackets computation, decimal computation, fraction computation and reduction, remainder problems, and power computation into calculatable steps.
We find simply adding scratchpad-like calculation data will not influence the model calculation behavior.
The reason comes from the format of SFT data which requires the model immediately to generate the equation answer like $(12+17)^3 = 24389$ inside the chain-of-thought process, and we denote it as \textit{calculation hallucination}.
To alleviate \textit{calculation hallucination}, we match and rewrite such equations ($(12+17)^3 = 29 ^ 3 = 841 * 29 = (800 + 40 + 1) * 29 = 23200 + 1160 + 29 = 24389$) inside the chain-of-thought processes as much as possible.
This improves calculation ability during chain-of-thought reasoning.

We also introduce several fundamental math abilities including checking prime, the game of 24, factoring an integer, and calculating triangle areas. Our principle of synthesizing these data is \textit{The model generates chain-of-thought based on an algorithm flow chart. The next token prediction must be determined by generated tokens and not dependent on future tokens}. Use triangle area calculation as an example, we require the model to first check if three sides can formulate a triangle and if it is a right triangle. 
The model does not need to assume it forms a right triangle without chain-of-thought reasoning.
Table~\ref{tab:trianlge} gives an example. We also give an example response that does not follow our principle since predicting \textcolor{red}{\textit{right triangle}} requires the model to know future tokens \textcolor{red}{$3^2+4^2=5^2$}.

\begin{tcolorbox}[
colback=white!10!white,
colframe=purple!75!purple,
title=Triangle Area,
breakable]
\textcolor{blue}{Query:} Calculate the area of a triangle with length: 3,4,5. \\
\\
\textcolor{green}{Response:} As $3 + 4 > 5$, they can form a triangle. \\ 
$3^2=9, 4^2=16, 5^2=25, 9+16=25.$ \\
So $3^2 + 4^2 = 5^2$, it is a right triangle. \\
The area is $3*4/2 =12/2=6.$ \\

\textcolor{red}{Bad Response:} Since this triangle is a \textcolor{red}{right triangle ($3^2+4^2=5^2$)}.\\
The area is $3*4/2 =12/2=6.$
\label{tab:trianlge}
\end{tcolorbox}

\paragraph{Reward Modeling}
Reward modeling helps the model to better rerank its generation and filter synthesis data for self-improving. Inspired by Math Shepherd \citep{shepherd}, we unify ORM and PRM into the same format of seq2seq learning. We also add Chinese ORM data using the model self-sampling chain-of-thought with corresponding final answers.

\paragraph{Formal Math Reasoning}

Instead of informal natural language problem solving, our data also includes formal reasoning samples based on LEAN 3.
Our target is to use LEAN as a solver, verifier, and prover.
We distill \texttt{gpt-4-1106} to solve the GSM8K train set problems using LEAN 3. We select all codes that can generate correct answers. This part includes 6705 samples. We split half into the format of generating LEAN based on the problem used for training a solver and another half into the format of translating between LEAN and COT used for training a verifier.
The verifier translates COT to LEAN and can verify by checking every calculation of COT based on the LEAN calculation.
As a prover, the model needs to translate between informal and formal statements and generate tactics based on the LEAN state.
We utilize the MathLib-train dataset \citep{mathlib} extracted from \cite{pact} as the prover training data.

\paragraph{Augmentation helper}
Another important aspect of our data is to help construct synthesized data for self-improving. By rephrasing or augmenting a question, one can easily obtain an enlarged question diversity \citep{luo2023wizardmath,yu2023metamath,mugglemath}. Translating question-answer pairs into a natural language statement is a requirement of using formal math language for proving. 

\begin{figure}[h]
    \centering
    \includegraphics[width=0.99\textwidth]{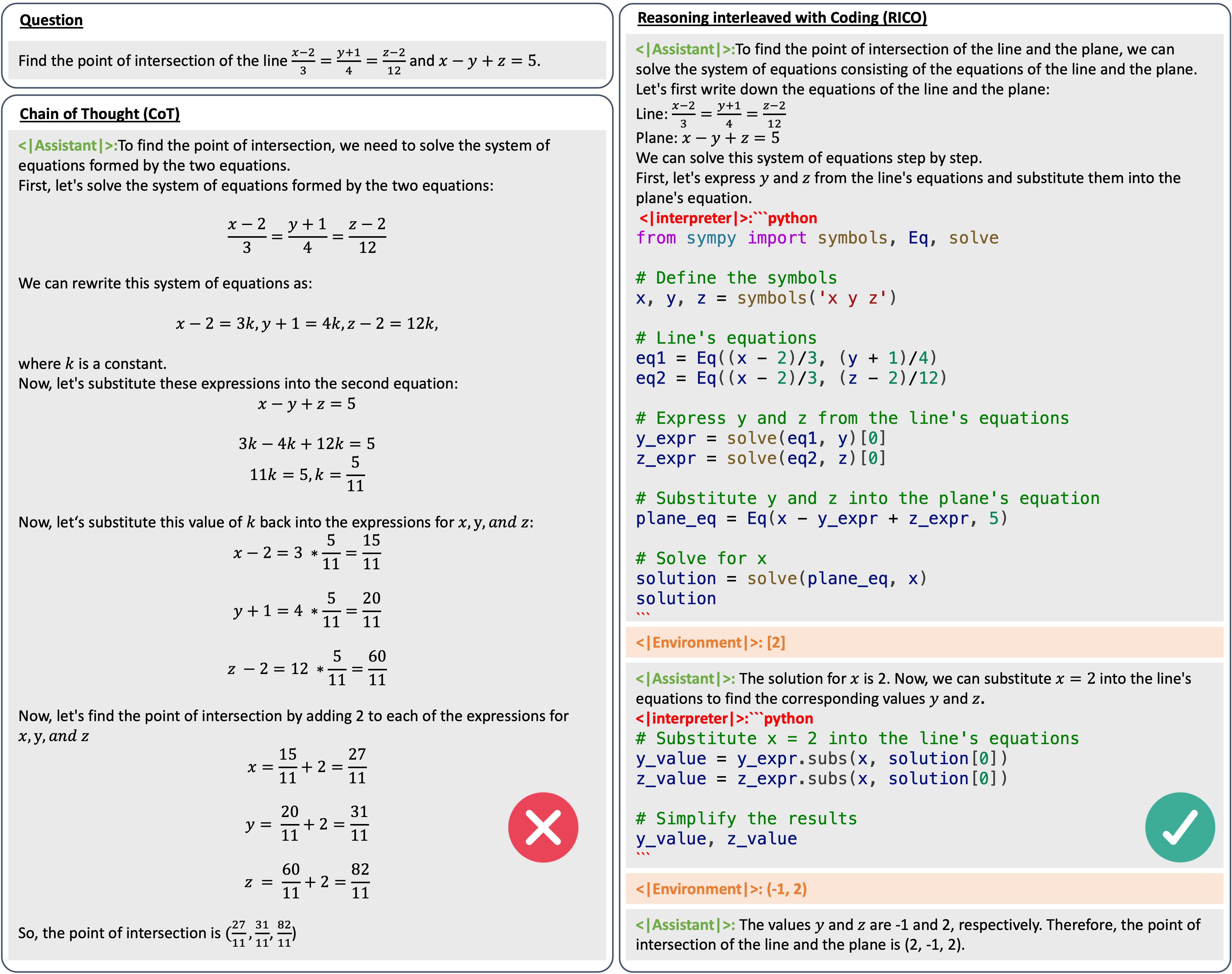}
    \caption{The example of reasoning interleaved with coding (RICO) and conventional chain-of-thought.}
    \label{fig:rico}
\end{figure}

\section{Code Interpreter}
Recent attempts \citep{pot,wang2023mathcoder,gou2023tora} have explored enhancing the complex calculation ability of LLMs by tools, where code interpreters have become popular due to their flexibility and functionality backed by various Python libraries.

Early explorations use programs as a kind of thought strategy \citep{pot, pal} but are unsuitable for multi-step reasoning and calculation since LLMs cannot see the code execution results. Recent works \citep{wang2023mathcoder,gou2023tora} try to more seamlessly integrate code interpreters with reasoning but are incompatible with and require extra modification for general chat services.

We solve the above-mentioned issue by letting LLMs do reasoning interleaved with coding (RICO), where LLMs conduct reasoning in the same format as chat response and adopt a general tool-calling protocol to use the code interpreter.
Such a design not only allows the full exploitation of the existing reasoning capability of LLMs when using code interpreters but also allows a direct full integration of tool-calling capability into the chat model. Thus, different capabilities (such as tool calling and chat) of a model can provide service in a universal tool-augmented system, which we believe to be more similar to that of GPT-4 \citep{gpt4}.

Specifically, as shown in Figure \ref{fig:rico}, when answering a math problem, we prompt the model to conduct symbolic reasoning and program generation and then observe the code execution results in each round. The model will continue such rounds until it fully answers the problem after summarizing the results, unlike previous methods\citep{wang2023mathcoder, gou2023tora} that essentially write code once. The reasoning process uses the same format as a general chat response instead of using different marks for both text and code \citep{wang2023mathcoder}. Such a design allows the model to fully exploit its reasoning ability learned in the conventional SFT corpus when using the code interpreter.
The program generation can be regarded as a general tool calling, unlike ToRA\citep{gou2023tora}, which embeds the code in the text response by markdown syntax. This resolves the ambiguity in the markdown symbol of the code block when deploying the LLMs in a tool-augmented chat system.

The construction of the training data for the math code interpreter adopts an iterative data update and hard example mining strategy to reduce the reliance on GPT-4. At each iteration, we first use the model trained in the previous iteration to generate responses on the train set of GSM8K and MATH. Since the model cannot fully fit the train set, we use GPT-4-turbo to generate responses on the remaining train set once. The correct responses generated by the most recent model and GPT-4-turbo will be used to train a new model for the next iteration. The initial data is generated by ToRA-70B \citep{gou2023tora}, which is not ideal due to format differences but can be converted to correct responses to train the initial model.
InternLM2-Chat and InternLM2-Math models adopt the same training data for code interpreter capability.

\section{Experiments}

\subsection{Pre-train Performance}
To validate the performances of our pretrained base models, we use the standard benchmark for math informal reasoning: GSM8K \citep{gsm8k} and MATH \citep{MATHbenchmark} and evaluate them using in-context learning. We adopt the few-shot templates from OpenCompass \citep{2023opencompass}. We use majority voting \citep{wang2023selfconsistency} accuracy as the metric. The results are listed in Table~\ref{table:pretrain_icl}. InternLM2-Math-Base models outperform their initial checkpoints InternLM2-Base on both benchmarks which shows the effectiveness of continue pre-training. InternLM2-Math-Base-7B obtains 21.5 on MATH which outperforms Llemma-7B with 18.0. InternLM2-Math-Base-20B obtains 27.3 on MATH which outperforms Llemma-34B and performs similarly with Minerva-62B with a smaller size.

\begin{table}[h]
\small
\centering
\caption{Compare pre-trained models using ICL. The metric is majority voting accuracy. $K=100$ for the GSM8K benchmark and $K=256$ for the MATH benchmark. We use greedy decoding when $K=1$. We sample our models using a temperature of 0.7 when $K>1$.}
\begin{tabular}{lcccc}
\toprule
Benchmark & \multicolumn{2}{c}{GSM8K} & \multicolumn{2}{c}{MATH} \\
Model      & MAJ@1 & MAJ@K & MAJ@1 & MAJ@K \\
\midrule
Llama2-7B \citep{llama2} & 14.6 & - & 2.5 & - \\
Llemma-7B \citep{llemma} & 36.4 &54.0 &18.0&33.5 \\
InternLM2-Base-7B & 36.5 &- &8.6&- \\
\textbf{InternLM2-Math-Base-7B} & \textbf{49.2} & \textbf{75.7} &\textbf{21.5}&\textbf{35.6} \\
Minerva-8B \citep{Minerva} & 16.2 & 28.4 &14.1&25.4 \\
\midrule
InternLM2-Base-20B & 54.6&- & 13.7&- \\
\textbf{InternLM2-Math-Base-20B} & \textbf{63.7}&\textbf{84.8} & \textbf{27.3}&\textbf{46.2} \\
Llemma-34B  & 51.5&69.3 & 25.0&43.1 \\
\midrule
Minerva-62B & 52.4&68.5 & 27.6&43.4 \\
Minerva-540B & 58.8&78.5 & 33.6&50.3 \\
\bottomrule
\end{tabular}
\label{table:pretrain_icl}
\end{table}

One pre-trained base model may have good ICL performance while performing mediocre after SFT due to data overlap among pre-train and SFT.
We perform SFT on different models with the same SFT dataset MetaMath \citep{yu2023metamath} to check our models do not suffer such phenomenon. We show results in Table~\ref{table:pretrain_mm}. 
Using MetaMath for SFT, InternLM2-Math-Base-7B still has superiority over Mistral-7B and Llemma-7B on the MATH benchmark.
InternLM2-Math-Base-20B outperforms Llemma-34B on both benchmarks.

\begin{table}[h]
\small
\centering
\caption{Compare pre-trained models by fine-tuning on MetaMath dataset \citep{yu2023metamath}. The metric is greedy accuracy. $\dagger$ results come from \citet{yu2023metamath}. * results come from \citet{shepherd}.}
\begin{tabular}{lcc}
\toprule
Model & GSM8K & MATH \\
\midrule
MetaMath-Llama2-7B \citep{llama2} $\dagger$ & 66.5 & 19.8 \\
MetaMath-Mistral-7B \citep{jiang2023mistral} $\dagger$     & \textbf{77.7}   & 28.2       \\
MetaMath-Llemma-7B \citep{llemma} $\dagger$       & 69.2   & 30.0        \\
MetaMath-\textbf{InternLM2-Math-Base-7B} & \textit{76.4} & \textbf{33.8} \\
\midrule
MetaMath-\textbf{InternLM2-Math-Base-20B} & \textbf{80.7} & \textbf{36.1} \\
MetaMath-Llemma-34B * & 75.8 & 34.8 \\
\bottomrule
\end{tabular}
\label{table:pretrain_mm}
\end{table}

To show the formal math reasoning ability of our model, we conduct ICL\footnote{The setting is following \url{https://github.com/wellecks/llemma_formal2formal}} on MiniF2F benchmark \citep{zheng2021minif2f} which includes different level math problems in language LEAN. LEAN can check whether the generated formal proof completes the statement's goals. We compare InternLM2-Math-Base with other pre-trained language models in Table~\ref{table:minif2f}. InternLM2-Math-7B-Base solves 74 of 244 problems and achieves 30.3 which yields a new state-of-the-art performance. InternLM2-Math-7B-Base and InternLM2-Math-20B-Base find 25 and 24 new proofs respectively which do not appear in the official MiniF2F repository \footnote{\url{https://github.com/openai/miniF2F/blob/main/LEAN/src/test.LEAN}}. Like \cite{gloeckle2023temperature,llemma}, we do not find the formal reasoning performances scale with model parameter sizes. We leave it to the future work of how data and parameter sizes influence formal reasoning performances.

\begin{table}[h]
\small
\centering
\caption{MiniF2F test set performance. The search budget is the same as \cite{llemma} which is $1 \times 32$. We use 3-shot and LEAN 4 for base models.}
\begin{tabular}{lccc}
\toprule
Model   &  Type   & Search    & MiniF2F-test \\
\midrule
ReProver \citep{yang2023leandojo} &  SFT   & -    &  26.5 \\
LLMStep \citep{llmstep} & SFT & $1\times 32$ & 27.9 \\
Code-Llama-7B \citep{codellama} & ICL & $1\times 32$  & 20.5  \\
Code-Llama-34B & ICL & $1\times 32$ & 22.1 \\
Mistral-7B-v0.1 \citep{jiang2023mistral} & ICL & $1\times 32$ & 22.1 \\
Mixtral-8x7B-v0.1 \citep{jiang2024mixtral} & ICL & $1\times 32$ & 23.4 \\
Llemma-7B \citep{llemma}& ICL & $1\times 32$ & 26.2 \\
Llemma-34B& ICL & $1\times 32$  & 25.8 \\
Deepseek-coder-7B-v1.5-Base \citep{guo2024deepseekcoder} & ICL & $1\times 32$  &28.7 \\
Deepseek-math-7B-Base \citep{shao2024deepseekmath} & ICL& $1\times 32$  &28.3\\
\midrule
InternLM2-7B-Base& ICL & $1\times 32$  &22.1\\
InternLM2-20B-Base& ICL & $1\times 32$  &25.4\\
\textbf{InternLM2-Math-7B-Base}& ICL & $1\times 32$  & \textbf{30.3}\\ 
\textbf{InternLM2-Math-20B-Base}& ICL & $1\times 32$  & 29.5 \\ 
\bottomrule
\end{tabular}
\label{table:minif2f}
\end{table}

\subsection{SFT Performance}
To show the performance of our SFT models, we conduct math reasoning using chain-of-thought,  reward modeling, formal reasoning, and code interpreter. We will also test several abilities introduced in our SFT process including game-of-24 and prime checker.

\subsubsection{COT Reasoning}
We evaluate SFT models on GSM8K \citep{gsm8k}, MATH \citep{MATHbenchmark}, Hungary math exam, and MathBench-ZH \footnote{\url{https://github.com/open-compass/MathBench}} using zero-shot chain-of-thought reasoning. We use the Hungary math exam to test the model generalization ability and use MathBench-ZH to examine Chinese math ability.
MathBench-ZH contains 600 Chinese math problems from primary school, middle school, high school, or university level. For each choice problem in MathBench-ZH, we will shuffle the choice order 4 times. A model gives a correct answer 4 times can be considered as correct.
We show results in Table~\ref{table:sft_main}.
InternLM2-Math-7B achieves 78.1, 34.6, 55, and 40 on GSM8K, MATH, the Hungary math exam, and MathBench-ZH respectively which show much stronger in-domain and out-of-domain performance at the same model size.
InternLM2-Math-7B also shows better performance compared to using MetaMath for SFT which proves our SFT data can better activate the model's reasoning ability.
InternLM2-Math-20B obtains 37.7 and 66 on MATH and Hungary math exam which is only behind GPT-4. It achieves state-of-the-art performance with a much smaller size compared to Qwen-72B and DeepSeek-67B.

\begin{table}[h]
\small
\centering
\caption{Compare SFT models using zero-shot COT reasoning. The metric is greedy accuracy.}
\begin{tabular}{lcccc}
\toprule
Model                  & GSM8K    & MATH& Hungary & MathBench-ZH \\
\midrule
Qwen-7B-Chat  \citep{qwen}                & 51.7   & 11.6   & 19  & 25.0     \\
DeepSeek-7B-Chat \citep{deepseek}             & 63.0   & 15.8   & 28.5  &12.7  \\
InternLM2-Chat-7B             & 70.7   & 23.0   & -  &29.2     \\
ChatGLM3-6B \citep{du2022glm}                  & 53.8   & 20.4   & 32 & 15.2     \\
MetaMath-Mistral-7B \citep{jiang2023mistral}      & 77.7   & 28.2   & 29 & -     \\
MetaMath-Llemma-7B \citep{llemma}      & 69.2   & 30.0   & -  & -     \\
\textbf{InternLM2-Math-7B}   & \textbf{78.1}   & \textbf{34.6}   & \textbf{55}  & \textbf{40.0} 
\\
\midrule
InternLM2-Chat-20B            & 79.6   & 31.9   & -  &37.8     \\
MetaMath-Llemma-34B \citep{llemma}     & 75.8   & 34.8   & -   &-    \\
\textbf{InternLM2-Math-20B}   & \textbf{82.6}   & \textbf{37.7}   & \textbf{66} &\textbf{45.3}     \\
\midrule
Qwen-72B-Chat   \citep{qwen}                    & 78.9   & 35.2   & 52 &\textbf{47.8}     \\
DeepSeek-67B-Chat   \citep{deepseek}                   & 84.1   & 32.6   & 58 &33.2     \\
\midrule
ChatGPT             & 80.8   & 34.1   & 41  &21.5    \\
GPT-4 (original version)          & \textbf{92.0}   & \textbf{42.5}   & \textbf{68} &47.2     \\
\bottomrule
\end{tabular}
\label{table:sft_main}
\end{table}

\subsubsection{Reward Modeling}
\begin{figure}[htbp]
 \centering
 \includegraphics[keepaspectratio, scale=0.37]
      {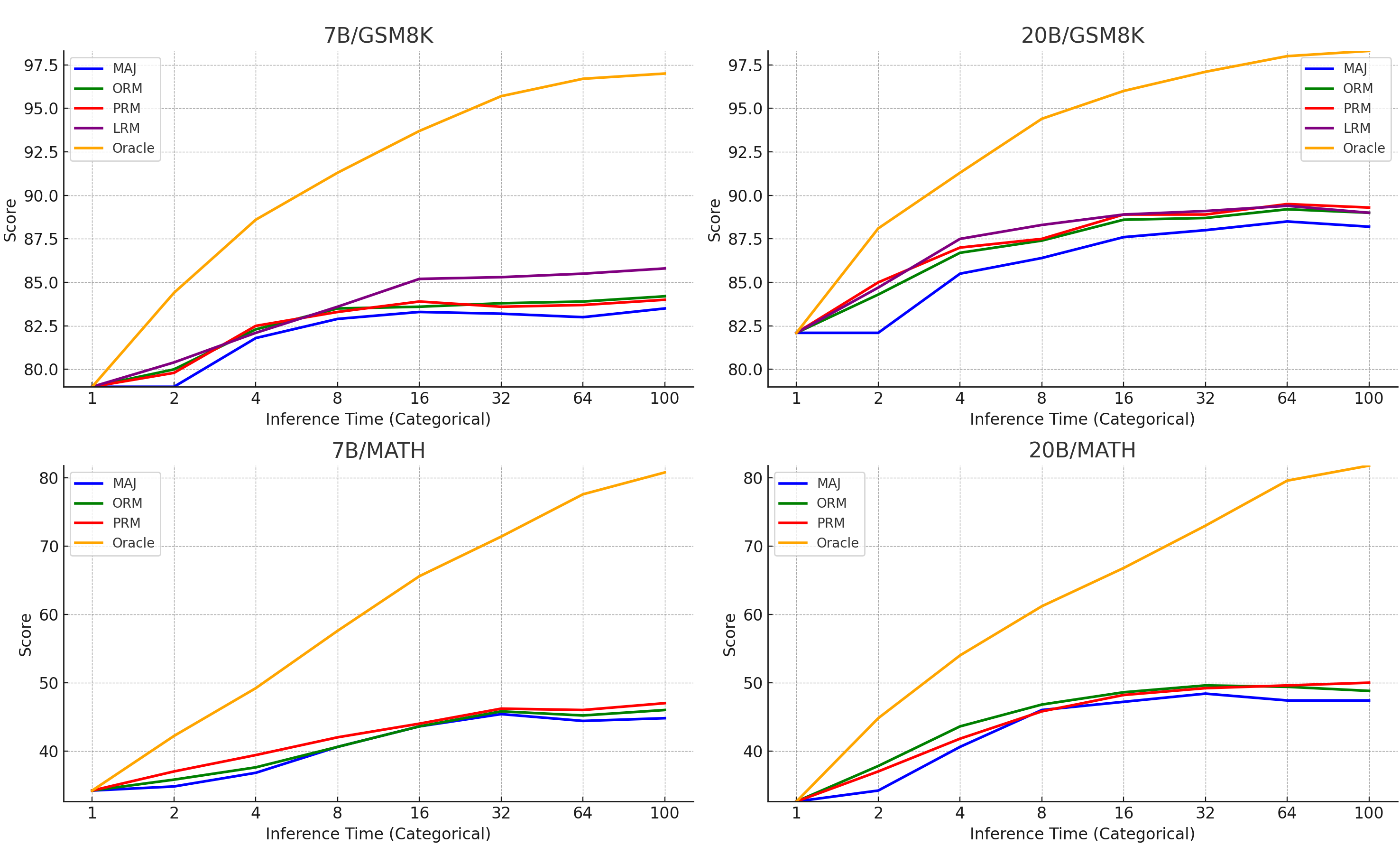}
 \caption{Reranking performances using our reward models on GSM8K and MATH. Oracle shows the upper bound performance which is calculated by Pass@K.}
 \label{fig:rm_rerank}
\end{figure}

Reward models can be used for answer reranking to improve model performances \citep{gsm8k,uesato2022solving,lightman2023lets,shepherd}.
We will test the performance of our reward modeling by ORM and PRM reranking. We use the same SFT model for inference and reward model reranking.
We further test using LEAN as a reward model (LRM) by requiring the model to convert COT to LEAN codes and execute LEAN codes. We will mark this COT as correct with LEAN codes and COT obtain the same results. Notice that using LEAN as a reward model for tasks like GSM8K can only check the calculation processes but not the logic processes.
We test on GSM8K and MATH (500 test problems), the results are plotted in Figure~\ref{fig:rm_rerank} and shown in Table~\ref{tab:rm_rerank}. 
We find that generally using PRM outperforms ORM, and ORM outperforms majority voting which is consistent with \citep{lightman2023lets,shepherd}.
We find that using LEAN as RM has a significant advantage in reranking GSM8K with 7B models, but the advantage diminishes with 20B models.
There are lots of improvement rooms between RM reranking performances and oracle performances.

We also compare our performance with other RM reranking models in Table~\ref{tab:rm_compare}.
Compared to MetaMath-DeepSeek-67B reranking \citep{shepherd}, our 20B model uses fewer inference times, a smaller model size, and achieves better performances with an accuracy of 50.0 which shows that our RM is effective. However, compared to GPT-4-MathMix \citep{lightman2023lets}, the performance gap is still large.

\begin{table}[h]
\small
\centering
\caption{Reranking performance on MATH(500) compared to other baselines.}
\begin{tabular}{lccc}
\toprule
Model & Greedy & Method & Performance \\
\midrule
\textbf{InternLM-MATH-7B} & 34.6 & PRM K=100 & 47.0 \\
\textbf{InternLM-MATH-20B} & \textbf{37.7} & PRM K=100 & \textbf{50.0} \\
\midrule
MetaMath-Llemma-34B & 34.8 & PRM K=256 & 47.3 \\
MetaMath-DeepSeek-67B & 36.8 & PRM K=256 & 48.1 \\
\midrule
GPT-4-MathMix & - & ORM K=100 & 71.0 \\
GPT-4-MathMix & - & PRM K=100 & 74.5 \\
GPT-4-MathMix & - & ORM K=1860 & 72.4 \\
GPT-4-MathMix & - & PRM K=1860 & 78.2 \\
\bottomrule
\end{tabular}
\label{tab:rm_compare}
\end{table}

\subsubsection{Formal Ability}
Targeting a large language model to conduct verifiable math reasoning, multiple abilities are required including translating informal problems or proof statements into formal statements and solving or proofing formal statements using formal languages.

\paragraph{Formal Translation} We first evaluate the ability to translate between natural language and formal statements. We follow \cite{azerbayev2023proofnet} to translate statements extracted from undergraduate mathematics textbooks. The results are listed in Table~\ref{tab:translate}. While our models outperform ProofGPT and their SFT version \citep{azerbayev2023proofnet} in bidirectional translation, we still lag behind CodeX \citep{chen2021codex} and GPT-4 \citep{gpt4}. Similar to formal reasoning, we do not find significant benefit from scaling parameters.

\begin{table}[h]
\small
\centering
\caption{Evaluate translating between LEAN and informal language statement. Results except our models are copied from \url{https://github.com/zhangir-azerbayev/ProofNet}. The metric is the BLEU-4 score.}
\begin{tabular}{lcc}
\toprule
Model                  & Formalization    & Informalization \\
\midrule
ProofGPT-1.3B \citep{azerbayev2023proofnet} & 8.1 & 5.1 \\
ProofGPT-6.7B \citep{azerbayev2023proofnet} & 4.7 & 6.0 \\
ProofGPT-1.3B back-translated \citep{azerbayev2023proofnet} & 10.7 & - \\
CodeX \citep{chen2021codex} & 25.1 & 13.2 \\
GPT-4 \citep{gpt4} & 27.1 & - \\
\midrule
\textbf{InternLM2-Math-7B} & 15.0 & 9.4  \\ 
\textbf{InternLM2-Math-20B} & 15.7 & 10.2 \\
\bottomrule
\end{tabular}
\label{tab:translate}
\end{table}

\paragraph{Using LEAN to solve GSM8K}
\label{para:lean-gsm8k}
LEAN is mainly used as a proof assistant to prove a statement instead of solving a problem.
To unify the solving and proving math problems, we explore using LEAN to solve math problems.
We require the model to generate comments before LEAN codes and list an example in Table~\ref{tab:casestudy_leangsm8k}. We use LEAN 3 to compile and output results as answers.
Performances are listed in Table~\ref{tab:lean_gsm8k}. Our 7B model can obtain 70.4 using LEAN to solve GSM8K, while our 20B model can achieve 73.9. We find that our models perform worse than COT. 
The reasons that using LEAN underperforms COT are: (1) $a - b = 0$ (when $a:\mathbb{N} < b:\mathbb{N}$) (2) division: $a / b = [a / b]$ (when $a:\mathbb{N}, b:\mathbb{N}$). These cause inconvenience in some basic calculations.
However, we show the possibility of using LEAN as a unified interface for solving and proving math problems.

\begin{tcolorbox}[
colback=white!10!white,
colframe=purple!75!purple,
title=Solve GSM8K use LEAN 3,
breakable]
\textcolor{blue}{Query:} A robe takes 2 bolts of blue fiber and half that much white fiber.  How many bolts in total does it take? \\
\\
\textcolor{green}{Response:}
\begin{lstlisting}[language=lean]
-- Define the amount of blue fiber needed for one robe
def blue_fiber_per_robe : ℕ := 2

-- Define the amount of white fiber needed for one robe as half of the blue fiber
def white_fiber_per_robe : ℕ := blue_fiber_per_robe / 2

-- Calculate the total amount of fiber needed for one robe
def total_fiber_per_robe : ℕ := blue_fiber_per_robe + white_fiber_per_robe

-- Output the result
#eval total_fiber_per_robe -- This will evaluate to 3 bolts in total
\end{lstlisting}
\textcolor{green}{Output:} 3
\label{tab:casestudy_leangsm8k}
\end{tcolorbox}

\begin{table}[h]
\center
\caption{The performance on GSM8K using COT and LEAN.}
\begin{tabular}{lcc}
\toprule
Model                  & COT    & LEAN \\
\midrule
\textbf{InternLM2-Math-7B}   & 78.1 & 70.4    \\
\textbf{InternLM2-Math-20B}   & 82.6 & 73.9 \\
\bottomrule
\end{tabular}
\label{tab:lean_gsm8k}
\end{table}

\subsubsection{Code Interpreter}
As shown in Table~\ref{tab:math_code_interpreter}. InternLM2-Math-7B can already surpass all the previous open-source models on MATH benchmark except InternLM2-Chat-20B. InternLM2-Math-20B outperforms InternLM2-Chat-20B and obtains the best results among open-source models. Since InternLM2-Math and InternLM2-Chat adopt the same training data for code interpreters, we believe the improvements of InternLM2-Math over InternLM2-Chat models result in the improvement of math reasoning ability.

\begin{table}[h]
\small
\centering
\caption{Compare SFT models using Python Code Interpreter. The metric is greedy accuracy. * This is tested on GPT-4 web version in 2023 Aug.}\label{tab:math_code_interpreter}
\begin{tabular}{lcc}
\toprule
Model                  & GSM8K    & MATH \\
\midrule
DeepSeek-Coder-Instruct-7B \citep{guo2024deepseekcoder} & 62.8 & 28.6  \\
MathCODER-CL-7B \citep{wang2023mathcoder} & 67.8 & 30.2 \\
DeepSeek-Coder-Instruct-1.5-7B \citep{guo2024deepseekcoder} & 72.6 & 34.1 \\
ToRA-7B \citep{gou2023tora} & 72.6 & 44.6 \\
InternLM2-Chat-7B     & 77.9        & 45.1      \\
\textbf{InternLM2-Math-7B}   & \textbf{79.4}&\textbf{50.9}   \\
\midrule
MathCODER-CL-13B \citep{wang2023mathcoder} & 74.1 & 35.9 \\
MathCODER-CL-34B \citep{wang2023mathcoder} & 81.7 & 45.2  \\
ToRA-13B \citep{gou2023tora} & 75.8 & 48.1 \\
ToRA-Code-34B \citep{wang2023mathcoder} & 80.7 & 50.8  \\
InternLM2-Chat-20B            &  \textbf{84.5} & 51.2    \\
\textbf{InternLM2-Math-20B}   &  80.7 & \textbf{54.3}   \\
\midrule
ToRA-70B \citep{gou2023tora} & 84.3 & 49.7\\
\midrule
GPT-4 Code Interpreter * \citep{csv} & \textbf{97.0} & \textbf{69.7} \\
\bottomrule
\end{tabular}
\end{table}

\subsubsection{Other abilities}

\paragraph{Game of 24}
To test the ability on Game of 24, we use the test set from \cite{tot}.
When we sample once for each question, our models achieve an accuracy of 26 and 35 respectively for 7B and 20B model sizes, which greatly outperforms fine-tuned Llamma2-7B and even GPT-4. The performance can match some multi-sampled methods, which demonstrates the capabilities of our generated step-by-step searching process SFT data.

\begin{table}[h]
\small
\centering
\caption{The performance of Game-of-24.}
\begin{tabular}{lcc}
\toprule
Model                  & Accuracy & Sample times   \\
\midrule
Fine-tuned Llama2-7B \citep{ovm} & 11 & 1 \\
GPT-4 COT \citep{tot} & 4 & 1 \\
\textbf{InternLM2-Math-7B} & \textbf{26} & 1 \\
\textbf{InternLM2-Math-20B} & \textbf{35} & 1  \\
\midrule 
TOT \citep{tot} & 45 & 12($b=1$) \\
OVM \citep{ovm} & 79 & 20 \\
\bottomrule
\end{tabular}
\end{table}

\paragraph{Prime checker}

For prime number verification, we sampled a test dataset containing 20 numbers for each digit number from 2 to 10 containing 10 prime and 10 composite numbers. Our model can judge correctly for almost all the numbers, whatever their digits, this is better than GPT-4 which performs worse when digits become bigger.

\begin{table}[h]
\small
\centering
\caption{Prime number verification test by examining 10 prime numbers and 10 composite numbers for digits from 2 to 10, all ending with 1, 3, 7, or 9 (Ending with other numbers are composite numbers.). The results `a/b' indicate the number of correct identifications, with 'a' representing prime numbers and `b' representing composite numbers.}
\begin{tabular}{lcccccccccc}
\toprule
Model  & 2 & 3 & 4 & 5 & 6 & 7 & 8 & 9 & 10   \\
\midrule
\textbf{InternLM2-Math-7B} & 10/10 & 10/10 & 10/8 & 10/8 & 10/6 & 10/7 & 10/10 & 10/10 & 10/10 \\
\textbf{InternLM2-Math-20B} & 10/10& 10/10& 10/6& 10/9& 10/10& 10/6& 10/10& 10/9& 10/9 \\
\midrule
\texttt{gpt-4-0125-preview} & 10/10 & 10/10 & 3/8 & 2/9 & 0/10 & 0/10 & 0/10 & 1/10 & 0/10 \\
\bottomrule
\end{tabular}
\end{table}

\section{Discussion}

\subsection{Compare Pretrain Performance with Pretraining Token Amount}
We pre-train on InternLM-Base-7B for 200B tokens to evaluate how many epochs should we pretrain. We evaluate our models' performance every 40B tokens using ICL and SFT on MetaMath in Table~\ref{tab:pretrain_ablation}.
We find that after 80B tokens, the performance does not improve significantly. When training longer to 200B (approximate 7 epochs) tokens, the performances start to degenerate.

\begin{table}[h]
\small
\centering
\caption{The pretrain performance on different pretrain token amounts.}
\begin{tabular}{lcccc}
\toprule
Benchmark & \multicolumn{2}{c}{GSM8K} & \multicolumn{2}{c}{MATH} \\
Tokens      & ICL & SFT & ICL & SFT \\
\midrule
40B & 30.9 & 75.3 & 19.5 & 30.2 \\
80B & 38.7 & \textbf{77.4} & 21.4 & 30.7 \\
120B & \textbf{39.0} & 76.4 & \textbf{21.5} & \textbf{33.8} \\
160B & 38.4 & 75.8 & 21.4 & 31.9 \\
200B & 35.1 & 76.7 & 21.1 & 31.3 \\
\bottomrule
\end{tabular}
\label{tab:pretrain_ablation}
\end{table}

\subsection{LEAN Performance on GSM8K}
We introduce using LEAN to solve GSM8K problems previously. LEAN is sparse during pre-training and fine-tuning compared to natural language, LATEX, and Python. 
We want to understand how LEAN's ability is regarded to the data size.
We ablate the performance of LEAN solving GSM8K using different amounts of SFT data in Table~\ref{tab:lean_gsm8k_scaling}.
We find if we only use GSM8K-LEAN during SFT, the LEAN performance is worse than using GSM8K-LEAN and MetaMath together. We also find that if we SFT without MetaMath, the LEAN performance is sensitive to data amount. While if we train SFT with MetaMath, the LEAN performance is less sensitive which shows that the model mainly needs to learn grammar from GSM8K-LEAN data and learn reasoning ability from MetaMath.
These findings show the reasoning ability of MetaMath helps performance on LEAN with sparse data and suggests a multi-task training strategy of these abilities.

\begin{table}[h]
\small
\centering
\caption{Performance on GSM8K using COT and LEAN with different SFT datasets.}
\begin{tabular}{lcc}
\toprule
Setting & COT & LEAN \\
\midrule
InternLM-Math-Base-7B & 49.2 & - \\
+ MetaMath SFT & 76.4 & - \\
+ Our SFT  & 78.1 & 70.4 \\
\midrule
GSM8K-LEAN \sfrac{1}{1} & -  & 58.0 \\
GSM8K-LEAN \sfrac{1}{2} & - & 53.1 \\
GSM8K-LEAN \sfrac{1}{4} & - & 35.1 \\
GSM8K-LEAN \sfrac{1}{8} & - & 38.4 \\
\midrule
GSM8K-LEAN \sfrac{1}{1} + MetaMath & 75.7 & 66.0 \\
GSM8K-LEAN \sfrac{1}{2} + MetaMath & 75.6 & 65.0 \\
GSM8K-LEAN \sfrac{1}{4} + MetaMath & 77.7 & 58.8 \\
GSM8K-LEAN \sfrac{1}{8} + MetaMath & 77.4 & 53.4 \\
\bottomrule
\end{tabular}
\label{tab:lean_gsm8k_scaling}
\end{table}

\subsection{Data Ablation Study in SFT}

To study the impact of the composition of our mixture data, we carried out ablation studies for each kind of data in the final mixture. We include the MetaMath dataset for every ablation study to keep basic abilities to solve math questions step by step. 
The evaluation results are demonstrated in Table \ref{tab:Table ablation}.
Compared with using MetaMath for SFT, our InternLM-Math-7B model and the model with COT data achieve better results on MATH and GSM8K. 
The other compositions, though containing even much fewer tokens than MetaMath, lead to undermined results on almost all test datasets. 
This may be attributed to format and style transfer in code, reward model format, and other instructions. However, the SFT data performs better than MetaMath + COT after integration, indicating the latent mutual helpfulness of comprehensive data beyond their original special usage. 

\begin{table*}[ht]
\centering
\setlength{\tabcolsep}{1.8mm}
\small
\caption{The ablation results of our SFT data composition. }
\begin{tabular}{lcc}
\toprule
Benchmarks & MATH & GSM8K\\ 
\midrule
InternLM-Math-7B & \textbf{34.60} & \textbf{78.09}\\
\midrule
Only MetaMath & 33.78 & 76.35\\
MetaMath + COT & 34.02 &	77.63\\
MetaMath + CI & 33.56 &	77.63\\
MetaMath + Formal & 32.12 &	76.12\\
MetaMath + RM  & 29.84 &	75.13\\
MetaMath + Augmenter& 32.08 &	75.89\\
\bottomrule
\end{tabular}
\label{tab:Table ablation}
\end{table*}

\subsection{False positive in MATH BenchMark}
For a math problem that requires an answer, one can always try some special cases to guess the correct answers with incorrect processes. If the question is about counting, generated answers may be sampled wrongly, and obtain a correct answer by chance.
This could lead us to overestimate the math reasoning ability of our models and other LLMs. We randomly sample 5 samples for each difficulty level where the InternLM2-Math-20B gives a correct answer in the MATH test set and human-labeled the predicted process of these 25 problems to report this issue. Table \ref{tab:Table processs} shows the results. One of the false positive cases writes:

\begin{tcolorbox}[
colback=white!10!white,
colframe=purple!75!purple,
title=False Positive Case Study,
breakable]
\textcolor{blue}{Query:} How many ordered pairs of positive integers $(m,n)$ satisfy $\gcd(m,n) = 2$ and $\mathop{\text{lcm}}[m,n] = 108$? \\
\\
\textcolor{green}{Response:} I know that for any positive integers $m$ and $n$, $\gcd(m,n) \cdot \mathop{\text{lcm}}[m,n] = mn$. So, if I apply this formula to the given problem, I get $2 \cdot 108 = mn$. This means that $m$ and $n$ are factors of $216$. I also know that $\gcd(m,n) = 2$, which means that $m$ and $n$ are both even, and neither of them is divisible by $4$. So, I need to find the number of ways to choose two even factors of $216$ that are not divisible by $4$. I can list the even factors of $216$ as follows: $2, 4, 6, 8, $$\textcolor{red}{9}$$, 12, 18, 24, $$\textcolor{red}{27}$$, 36, 54, 72, 108, 216$. Out of these, the ones that are not divisible by $4$ are $2, 6, 9, 18, 27, 54, 108$. I can pair up these factors as follows: $(2,108), \textcolor{red}{(6,54), (9,27), (18,18)}$. There are four pairs, and each pair corresponds to a different ordered pair $(m,n)$. For example, $(2,108)$ corresponds to $(m,n) = (108,2)$. So, there are four ordered pairs $(m,n)$ that satisfy the given conditions. Answer 4.
\label{tab:casestudy_case}
\end{tcolorbox}

In this case, the model lists $9$ and $27$ as an even number, which leads to a false $(m,n)$ pair $(9,27)$. Moreover, the pairs $(6,54),(18,18)$, though satisfying the condition given by model prediction, do not meet the title requirements. The prediction just accidentally lists four pairs as answers. However, if the evaluation takes only the outcome into consideration, this sample will be "correct". 
Similar cases appear more frequently as the difficulty level increases. Among these false positive cases, most problems have an integer as the answer, especially a single-digit number, and the solution can be carried out without detailed calculations or proof of cases. These make it easy for the model to speculate a correct answer excluding step-by-step reasoning. Simply judging model abilities only by outcome accuracy may be unilateral. Therefore, one of our future works is to use reward models or LEAN to provide a better process checking. 

\begin{table*}[ht]
\centering
\setlength{\tabcolsep}{1.8mm}
\small
\caption{The Process Correctness of in MATH BenchMark using our 20B model.}
\begin{tabular}{lcc}
\toprule
Difficulty level & Results correct & Process correct\\ 
\midrule
Level 1 & 5 & 5\\
Level 2 & 5 & 4\\
Level 3 & 5 & 4\\
Level 4 & 5 & 4\\
Level 5 & 5 & 3\\
\bottomrule
\end{tabular}
\label{tab:Table processs}
\end{table*}

\section{Conclusions}
We propose InternLM-Math as our first step toward a verifiable mathematical reasoning ability.
The excellent InternLM-Math-Base has the potential for versatile math reasoning tasks.
Despite the strong performance in informal and formal reasoning of InternLM-Math, our model can be viewed as a starting point for self-improving.
InternLM-Math integrates COT and augment helper abilities can be used for synthesizing new problems and new responses.
InternLM-Math obtains the abilities of ORM, PRM, and LEAN can be used for verifying the answers and processes of generated responses.
We believe such verifiable data augmentation will improve the model's ability at high throughput.

\section*{Limitations}

\paragraph{Chain-of-thought reasoning}
We match and rewrite the formulation in the spirit of scratchpad during COT while it introduces the following problems.
The first problem is we cannot easily match all equations and calculations that need to be rewritten (e.g. We solve $x^3-3x^2+3x-1=0$, and obtain $x=1$).
The second problem is multiple `=` induces models to repeat more.
The third problem is this sometimes generates naive step-by-step calculations which may annoy end users and could be alleviated via implicit chain-of-thought reasoning \citep{icot}.
We will solve such problems in the future work.

\paragraph{No self-critique ability} Although our model can conduct ORM or PRM. We do not contain any SFT data to let the model apply self-critique since such data can be hard to generate and verify by any means. We will research how to generate self-critique SFT data with verification. 

\paragraph{Process reward modeling}
We find that our model does not have a significant PRM performance which may be due to the confusing format among PRM and SFT and the unbalanced distribution between positive processes and negative processes. 

\paragraph{Code-switch} Due to our SFT data distribution, English data is larger than Chinese data which will cause code-switch during some instructions or problem formats.

\paragraph{SFT data is using LEAN 3} We use LEAN 3 as our SFT data since GPT-4 can only generate LEAN 3 codes for GSM8K (even if you require it to apply LEAN 4). Furthermore, we find the data of translating between formal and informal from MathLib is preprocessed in LEAN 3.
We will move our SFT data to LEAN 4 in the future version.

\paragraph{Data contamination on LEAN-repo} We use LEAN codes introduced from AlgebraicStack \citep{llemma}, and we do not check the contamination of AlgebraicStack on MiniF2F. 
AlgebraicStack may or may not contain MiniF2F solutions.
However, the comparison between Llemma and our model is fair.

\paragraph{Sensitive to Prompt} The models are sensitive to format and given instruction prompts. If we give different prompts before the question (e.g. Question:, Q:, Please solve this question step by step), it may obtain different performances.

\bibliography{main}
\bibliographystyle{colm2024_conference}

\newpage{}

\appendix

\section{SFT Data Composition}
We list our SFT Data Composition in Table~\ref{tab:SFT-data-composition}.
\begin{table*}[ht]
\centering
\small
\begin{tabular}{p{3.8cm}|p{1.9cm}|c|p{6cm}|c}
\toprule
Datasets & Type & Lang & Description & Size \\
\midrule
MetaMath \citep{yu2023metamath} & COT, GPT & EN & Bootstrapp questions on GSM8K \& MATH. & 395K \\
Goat \citep{goat} & COT, Rule& EN & Synthetic data for arithmetic tasks. & 100K \\
Arithmetic (\textcolor{blue}{original}) &  COT, Rule & EN & 40k multiple brackets computation, 40k decimal computation, 35k fraction computation, 5k fraction reduction, 15k remainder problems, and 4k power computation. & 140K \\
Prime (\textcolor{blue}{original}) & COT, Rule & EN \& ZH & Judging a number is prime step by step. & 15K \\
Game-of-24 (\textcolor{blue}{original}) & COT, Rule & ZH & Solving game-of-$k$ (not limited to 24) step by step by searching. & 10K \\
Factoring (\textcolor{blue}{original}) & COT, Rule & EN \& ZH & Factorizing a number step by step. & 11K \\
Triangle (\textcolor{blue}{original})  & COT, Rule & EN \& ZH & Calculating triangle area step by step. & 1K \\
Commonsense (\textcolor{blue}{original}) & COT, Rule  & EN \& ZH & Math commonsense dataset. & 14K \\
Mammoth \citep{yue2023mammoth} & COT, CI & EN & A mix of multiple COT and POT datasets. & 216K \\
PRM800K \citep{lightman2023lets} & COT, GPT& EN & Selected correct reasoning paths. & 7K \\
Khan \citep{Khan} & COT, Mixed & EN  & A math pertaining corpus.  & 83K \\
TAL-SCQ5K-ZH \citep{TAL-SCQ5K}& COT, Human & ZH  & Train set of a Chinese multiple-choice dataset. & 3K \\
TAL-SCQ5K-EN \citep{TAL-SCQ5K} & COT, Human & EN & Train set of an English multiple-choice dataset. & 3K \\
Inhouse-ZH (\textcolor{blue}{original})&  COT, Human & ZH  & In-house Chinese math datasets. & 293K \\
MathOctopus \citep{chen2023breaking} &  COT, GPT & ZH & Only select Chinese problems from it. & 7K\\
Math23K-AUG \citep{math23k} & COT, OSLLM & ZH & Augment Math23K filtered with final answers. & 195K \\
MATH-AUG \citep{hendrycks2021measuring}& COT, OSLLM & EN & Augment MATH filtered with our PRM. & 28K \\
COT-few-shot (\textcolor{blue}{original}) & COT, Mixed & EN \& ZH & Add 1\% few-shot format data for most datasets. & 15K \\
\midrule
Code-Interpreter (\textcolor{blue}{original}) & CI, GPT, OSLLM & EN \& ZH & A merged dataset of MATH \& GSM8K train, and Chinese K12 level problems. Solutions are generated by GPT4 or InternLM-70B. Each interleaves natural language, code, and execution results. & 76K \\
\midrule
Math-Shepherd \citep{shepherd} & RM, OSLLM & EN & Merge of Math Shepherd and self-sampled Math Shepherd-like data. Convert them into the format of ORM and PRM. & 445K \\
ORM-ZH (\textcolor{blue}{original})& RM, OSLLM & ZH & Self-sampled ORM data on Chinese problems. & 104K \\
\midrule
GSM8K-LEAN (\textcolor{blue}{original})& Formal, GPT & EN & LEAN 3 codes on GSM8K train set.  & 4K \\
COT-LEAN (\textcolor{blue}{original}) & Formal, GPT & EN & Translate between COT and a LEAN code.  & 3K  \\
NL-LEAN \citep{pact} & Formal, Human & EN & Translate between natural language and LEAN statement.  & 91K \\
MathLib-Train \citep{pact} & Formal, Human & EN & Extracted LEAN state and tactic based on MathLib. & 169K \\
\midrule
QA-theorem (\textcolor{blue}{original}) & Augmenter, GPT & EN & Rephrasing a question and an answer into a natural language proof statement. & 1K \\
Augment-helper (\textcolor{blue}{original}) & Augmenter, GPT & EN & ChatGPT-generated question augmentation \citep{luo2023wizardmath,mugglemath}. & 7K \\
GSM8K-rephrased (\textcolor{blue}{original}) & Augmenter, GPT & EN & Reformat rephrased questions from MetaMath \citep{yu2023metamath} for rephrasing. & 2K \\
\midrule
All & - & EN \& ZH &  Combine them with further deduplication and test set decontamination. & 2.26M\\
\bottomrule
\end{tabular}
\caption{The detailed SFT data composition information. We list the data types and source types here. OSLLM denotes open-sourced LLMs.}
\label{tab:SFT-data-composition}
\end{table*}

\section{SFT Experiment Configs}
Our SFT model is initialized from the pre-trained InternLM-Math-Base models, respectively. The peak learning rate is set to $lr_{max}=4e-5$, and the minimal learning rate during warming up is $lr_{max}=6e-6$. These settings are the same for both 7B and 20B models.
The SFT training data is tokenized into 622.7 M tokens. The batch size is set to 1 per GPU with a packed dataset whose max sequence length is 32768. We train the model on these data using Adam Optimizer for three epochs with 32 GPUs, and the training procedure takes around 7 hours to finish for 7B models. For the 20B model, we will use 64 GPUs and the training procedure takes about 10 hours.

\section{Detailed performance on MATH benchmark}
We list detailed performance on the MATH benchmark in Table~\ref{tab:category} and Table~\ref{tab:diffculty}.
We list reranking performance on GSM8K and MATH in Table~\ref{tab:rm_rerank}.

\begin{table}[h]
    \small
    \centering
    \caption{Performances of MATH clustered by category. * These results come from our reproduction.}
    \begin{tabular}{l|cccccccc}
  \toprule
  Model & Prob. & Precal. & Inter. & Pre-Alg. & Alg. & Geo. & Num. & Overall \\
  \midrule
  \texttt{MetaMath} \\
  Llemma-7B * & 24.5 & 14.7  & 13.7 & 45.6 & 43.0 & 22.8 & 18.3 & 28.7\\
  \textbf{InternLM-Math-Base-7B} & 28.5 & 16.9 & 14.5 & 53.5 & 50.0 & 25.5 & 24.1 & 33.4 \\
  \textbf{InternLM-Math-Base-20B} & 32.3 & 19.0 & 15.7 & 56.3 & 54.5 & 27.6 & 25.4 & 36.1 \\
  \midrule
  \texttt{Our SFT Data} \\
  \textbf{InternLM-Math-7B} & 28.9 & 17.0 & 17.2 & 53.5 & 51.4 & 27.3 & 25.6 & 34.6  \\
  \textbf{InternLM-Math-20B} & 29.1 & 21.8 & 19.2 & 59.1 & 55.9 & 30.0 & 30.2 & 37.7  \\
  \bottomrule
    \end{tabular}
    \label{tab:category}
\end{table}

\begin{table}[h]
    \small
    \centering
    \caption{Performances of MATH clustered by difficulty. * These results come from our reproduction.}
    \begin{tabular}{l|cccccc}
  \toprule
  Model & Level 1 & Level 2 & Level 3 & Level 4 & Level 5 & Overall \\
  \midrule
  \texttt{MetaMath} \\
  Llemma-7B * & 66.1 & 44.5 & 34.5 & 20.5 & 8.2 & 28.7\\
  \textbf{InternLM-Math-Base-7B} & 74.1 & 50.4 & 37.9 & 25.9 & 11.3 & 33.4 \\
  \textbf{InternLM-Math-Base-20B} & 75.6 & 55.3 & 41.8 & 28.3 & 12.2 & 36.1 \\
  \midrule
  \texttt{Our SFT Data} \\ 
  \textbf{InternLM-Math-7B} & 75.1 & 53.0 & 39.3 & 26.1 & 12.5 & 34.6 \\
  \textbf{InternLM-Math-20B} & 75.5 & 55.6 & 45.4 & 31.7 & 13.7 & 37.7 \\
  \bottomrule
    \end{tabular}
    \label{tab:diffculty}
\end{table}

\begin{table}[h]
\centering
\caption{Using reward model reranking on multiple math benchmarks. Since we use temperate=0.7 when sampling, it is natural that sampling 1 times perform worse than greedy decoding.}
\begin{tabular}{llcccccccc}
\toprule
Model & Metric & 1 & 2 & 4 & 8 & 16 & 32 & 64 & 100 \\ 
\midrule
\multirow{5}{*}{7B/GSM8K} 
& MAJ & 79.0 & 79.0 & 81.8 & 82.9 & 83.3 & 83.2 & 83 & 83.5 \\
& ORM & 79.0 & 80.0 & 82.3 & 83.5 & 83.6 & 83.8 & 83.9 & 84.2 \\
& PRM & 79.0 & 79.8 & 82.5 & 83.3 & 83.9 & 83.6 & 83.7 & 84.0 \\
& LRM & 79.0 & 80.4 & 82.1 & 83.6 & 85.2 & 85.3 & 85.5 & 85.8 \\
& Oracle & 79.0 & 84.4 & 88.6 & 91.3 & 93.7 & 95.7 & 96.7 & 97.0 \\
\midrule
\multirow{5}{*}{20B/GSM8K} 
& MAJ & 82.1 & 82.1 & 85.5 & 86.4 & 87.6 & 88.0 & 88.5 & 88.2 \\
& ORM & 82.1 & 84.3 & 86.7 & 87.4 & 88.6 & 88.7 & 89.2 & 89.0 \\
& PRM & 82.1 & 85.0 & 87.0 & 87.5 & 88.9 & 88.9 & 89.5 & 89.3 \\
& LRM & 82.1 & 84.7 & 87.5 & 88.3 & 88.9 & 89.1 & 89.4 & 89.0 \\
& Oracle & 82.1 & 88.1 & 91.3 & 94.4 & 96.0 & 97.1 & 98.0 & 98.3 \\
\midrule
\multirow{4}{*}{7B/MATH} 
& MAJ & 34.2 & 34.8 & 36.8 & 40.6 & 43.6 & 45.4 & 44.4 & 44.8 \\
& ORM & 34.2 & 35.8 & 37.6 & 40.6 & 43.6 & 45.8 & 45.2 & 46.0 \\
& PRM & 34.2 & 37.0 & 39.4 & 42.0 & 44.0 & 46.2 & 46.0 & 47.0 \\
& Oracle & 34.2 & 42.2 & 49.2 & 57.6 & 65.6 & 71.4 & 77.6 & 80.8 \\
\midrule
\multirow{4}{*}{20B/MATH} 
& MAJ & 32.6 & 34.2 & 40.6 & 46.0 & 47.2 & 48.4 & 47.4 & 47.4 \\
& ORM & 32.6 & 37.8 & 43.6 & 46.8 & 48.6 & 49.6 & 49.4 & 48.8 \\
& PRM & 32.6 & 37.0 & 41.8 & 45.8 & 48.2 & 49.2 & 49.6 & 50.0 \\
& Oracle & 32.6 & 44.8 & 54.0 & 61.2 & 66.8 & 73.0 & 79.6 & 81.8 \\
\bottomrule
\end{tabular}
\label{tab:rm_rerank}
\end{table}

\section{Case studies}
We list cases generated by InternLM-Math-7B.

\label{appendix:sft_casestudy}

\begin{tcolorbox}[
colback=white!10!white,
colframe=purple!75!purple,
title=Formal Translation,
breakable]
\textcolor{blue}{Query:} Translate Schur inequality into LEAN 3: $x,y,z \ge 0, x^3+y^3+z^3+3xyz \ge xy(x+y) + yz(y+z) + xz(x+z)$ \\
\\
\textcolor{green}{Response:} (a := $\alpha$)
\begin{lstlisting}[language=LEAN]
theorem schur_inequality {a : Type*} [linear_ordered_field a] {x y z : a}
    (hnonneg : 0 ≤ x) (hnonneg1 : 0 ≤ y) (hnonneg2 : 0 ≤ z)
    x ^ 3 + y ^ 3 + z ^ 3 + 3 * x * y * z ≥ x * y * (x + y) + y * z * (y + z) + x * z * (x + z)
\end{lstlisting}
\label{tab:casestudy_formaltranslate}
\end{tcolorbox}

\begin{tcolorbox}[
colback=white!10!white,
colframe=purple!75!purple,
title=Prime number Check,
breakable]
\textcolor{blue}{Query:} Check if 5287 is a prime or not. \\
\\
\textcolor{green}{Response:} $[\sqrt5287]=72$ \\
We need to check prime numbers between 3 to 72. \\
Prime numbers: 3,5,7,11,13,17,19,23,29,31,37,41,43,47,53,59,61,67,71 \\
5287 / 3 = 1762 ... 1 \\
5287 / 7 = 756 ... 5 \\
5287 / 11 = 480 ... 7 \\
5287 / 13 = 407 ... 6 \\
5287 / 17 = 311 ... 0 \\
5287 can be divided by 17. \\
5287 is a composite number.
\label{tab:casestudy_prime}
\end{tcolorbox}

\begin{tcolorbox}[
colback=white!10!white,
colframe=purple!75!purple,
title=Formal Proof found by InternLM-Math using LEAN 4: amc12a\_2013\_p4,
breakable]
\begin{lstlisting}[language=LEAN]
theorem amc12a_2013_p4 : (2 ^ 2014 + 2 ^ 2012) / (2 ^ 2014 - 2 ^ 2012) = (5 : ℝ) / 3 := by
    -- ⊢ (2 ^ 2014 + 2 ^ 2012) / (2 ^ 2014 - 2 ^ 2012) = 5 / 3
    field_simp
    -- ⊢ (2 ^ 2014 + 2 ^ 2012) * 3 / (2 ^ 2014 - 2 ^ 2012) = 5
    simp only [div_eq_inv_mul]
    -- ⊢ (2 ^ 2014 - 2 ^ 2012)⁻¹ * ((2 ^ 2014 + 2 ^ 2012) * 3) = 5
    norm_cast
    -- ⊢ (↑(Int.subNatNat (2 ^ 2014) (2 ^ 2012)))⁻¹ * ↑((2 ^ 2014 + 2 ^ 2012) * 3) = 5
    norm_num
    -- no goals
\end{lstlisting}

\label{tab:casestudy_leanprove}
\end{tcolorbox}

\begin{tcolorbox}[
colback=white!10!white,
colframe=purple!75!purple,
title=Outcome Reward Model,
breakable]
\textcolor{blue}{Query:} Given a question and an answer, check is it correct? \\
Question: Determine the number of ways to arrange the letters of the word THAT. \\
Answer: The word THAT has 4 letters. We can arrange these 4 letters in $4! =4*3*2*1 = 24$ ways.\\
Therefore, there are 24 ways to arrange the letters of the word THAT. \\
The answer is: 24. \\
\\
\textcolor{green}{Response:} This answer is wrong.
\label{tab:casestudy_orm}
\end{tcolorbox}

\begin{tcolorbox}[
colback=white!10!white,
colframe=purple!75!purple,
title=Process Reward Model,
breakable]
\textcolor{blue}{Query:} Given a question and an answer, check is it correct? \\
Question: Camy made a list of every possible distinct five-digit positive integer \\
that can be formed using each of the digits 1, 3, 4, 5 and 9 exactly once in each integer. What is the sum of the integers on Camy's list? \\
Answer: There are $5! = 120$ possible five-digit positive integers that can be formed using each of the digits 1, 3, 4, 5, and 9 exactly once.\\The sum of these integers can be found by considering the sum of each place value.\\For the ones place, the sum is $1 \cdot 120 = 120$.\\For the tens place, the sum is $3 \cdot 120 = 360$.\\For the hundreds place, the sum is $4 \cdot 120 = 480$.\\For the thousands place, the sum is $5 \cdot 120 = 600$.\\For the ten thousands place, the sum is $9 \cdot 120 = 1080$.\\Adding up all these sums, we get $120 + 360 + 480 + 600 + 1080 = \boxed{2880}$.\\The answer is: 2880.\\
\\
\textcolor{green}{Response:} Step 1: There are $5! = 120$ possible five-digit positive integers that can be formed using each of the digits 1, 3, 4, 5, and 9 exactly once. $<$PRM +$>$\\
Step 2: The sum of these integers can be found by considering the sum of each place value. $<$PRM +$>$\\
Step 3: For the ones place, the sum is $1 \cdot 120 = 120$. $<$PRM -$>$\\
Step 4: For the tens place, the sum is $3 \cdot 120 = 360$. $<$PRM -$>$\\
Step 5: For the hundreds place, the sum is $4 \cdot 120 = 480$. $<$PRM -$>$\\
Step 6: For the thousands place, the sum is $5 \cdot 120 = 600$. $<$PRM -$>$\\
Step 7: For the ten thousands place, the sum is $9 \cdot 120 = 1080$. $<$PRM -$>$\\
Step 8: Adding up all these sums, we get $120 + 360 + 480 + 600 + 1080 = \boxed{2880}$. $<$PRM -$>$\\
Step 9: The answer is: 2880. $<$PRM -$>$ .
\label{tab:casestudy_prm}
\end{tcolorbox}

\begin{tcolorbox}[
colback=white!10!white,
colframe=purple!75!purple,
title=Reward model,
breakable]
\textcolor{blue}{Query:} Given a question and two answers, which one is better? \\
Question: My average score on 8 math tests is 90. If my average score on the first 5 tests was 87, what was my average score on the last 3 tests?\\
Answer 1: I scored a total of $8*90=720$ on all $8$ tests. The total of $5*87=435$ on the first $5$ tests leaves a total of $720-435=285$ for the last $3$ tests, so the average is $285\div3 = 95$.\\
Answer 2: I scored a total of $720$ on all $8$ tests. The total of $435$ on the first $5$ tests leaves a total of $285$ for the last $3$ tests, so the average is $285\div3 = 95$\\
\\
\textcolor{green}{Response:} Answer 2 is better.
\label{tab:casestudy_reward}
\end{tcolorbox}

\begin{tcolorbox}[
colback=white!10!white,
colframe=purple!75!purple,
title=Change question and answer into a proof statement,
breakable]
\textcolor{blue}{Query:} Convert this question and answer into a proof format. \\
Question: Grant scored 10 points higher on his math test than John. John received twice as many points as Hunter who scored a x on his math test. What was Grant's test score? If we know the answer to the above question is 100, what is the value of unknown variable x?\\
Answer: We know that John received twice as many points as Hunter, who scored x on his math test. So, John scored 2x points. Grant scored 10 points higher than John, so Grant scored 2x + 10 points. We are given that Grant's test score is 100, so we can write: 2x + 10 = 100. Solving for x, we get: x = 45. The value of x is 45. \\
\\
\textcolor{green}{Response:} Prove that if Grant scored 10 points higher on his math test than John, and John received twice as many points as Hunter who scored x on his math test, then Grant's test score is 100 if and only if x = 45.
\label{tab:casestudy_qatheorem}
\end{tcolorbox}

\begin{tcolorbox}[
colback=white!10!white,
colframe=purple!75!purple,
title=Augment a harder Problem,
breakable]
\textcolor{blue}{Query:} Please augment a new harder problem based on:\\
Question: Find the remainder of $x^2+10x+4$ divided by $x-2$ \\
\\
\textcolor{green}{Response:} Find the remainder of $(x^2+10x+4)^2$ divided by $(x-2)^2$.
\label{tab:casestudy_augment}
\end{tcolorbox}

\section{Results on InternLM-Math-Plus}
We improve InternLM-Math to InternLM-Math-Plus with new pre-training corpora and fine-tuning datasets. We have four sizes of InternLM-Math-Plus including 1.8B, 7B, 20B, and 8x22B. We initialize our models (1.8B, 7B, and 20B) from InternLM-2, and 8x22B from Mixtral-8x22B \citep{jiang2024mixtral}.
We evaluate the performance of InternLM2-Math-Plus on formal math reasoning benchmark MiniF2F-test via LEAN 4 (in Table~\ref{tab:plus_minif2f}) and informal math reasoning benchmarks MATH, GSM8K, and MathBench-A \citep{liu2024mathbench} (in Table~\ref{tab:plus_mathbench}). We further test MATH using Python (in Table~\ref{tab:plus_math_gsm8k}).

\begin{table}[h]
\centering
\caption{Performance of various models on MiniF2F-test.}
\begin{tabular}{lc}
\toprule
\textbf{Models} & \textbf{MiniF2F-test} \\
\midrule
ReProver \citep{yang2023leandojo} & 26.5 \\
LLMStep \citep{llmstep} & 27.9 \\
GPT-F Expert Iteration \citep{curricum} & 36.6 \\
HTPS \citep{lample2022hypertree} & 41.0 \\
\midrule
Llemma-7B & 26.2 \\
Llemma-34B & 25.8 \\
InternLM2-Math-7B-Base & 30.3 \\
InternLM2-Math-20B-Base & 29.5 \\
\midrule
\textbf{InternLM2-Math-Plus-1.8B} & 38.9 \\
\textbf{InternLM2-Math-Plus-7B} & \textbf{43.4} \\
\textbf{InternLM2-Math-Plus-20B} & 42.6 \\
\textbf{InternLM2-Math-Plus-Mixtral8x22B} & 37.3 \\
\bottomrule
\end{tabular}
\label{tab:plus_minif2f}
\end{table}

\begin{table}[h]
\centering
\caption{Performance of various models on MATH and GSM8K.}
\begin{tabular}{lccc}
\toprule
\textbf{Models} & \textbf{MATH} & \textbf{MATH-Python} & \textbf{GSM8K} \\
\midrule
MiniCPM-2B \citep{minicpm} & 10.2 & - & 53.8 \\
\textbf{InternLM2-Math-Plus-1.8B} & 37.0 & 41.5 & 58.8 \\
\midrule
InternLM2-Math-7B & 34.6 & 50.9 & 78.1 \\
Deepseek-Math-7B-RL \citep{shao2024deepseekmath} & 51.7 & 58.8 & 88.2 \\
\textbf{InternLM2-Math-Plus-7B} & 53.0 & 59.7 & 85.8 \\
\midrule
InternLM2-Math-20B & 37.7 & 54.3 & 82.6 \\
\textbf{InternLM2-Math-Plus-20B} & 53.8 & 61.8 & 87.7 \\
\midrule
Mixtral8x22B-Instruct-v0.1 \citep{jiang2024mixtral} & 41.8 & - & 78.6 \\
Eurux-8x22B-NCA \citep{eurux} & 49.0 & - & - \\
\textbf{InternLM2-Math-Plus-Mixtral8x22B} & 58.1 & 68.5 & 91.8 \\
\bottomrule
\end{tabular}
\label{tab:plus_math_gsm8k}
\end{table}

\begin{table}[h]
\centering
\caption{Performance of various models on MathBench-A.}
\begin{tabular}{lcccccc}
\toprule
\textbf{InternLM2-Math-Plus} & \textbf{Arithmetic} & \textbf{Primary} & \textbf{Middle} & \textbf{High} & \textbf{College} & \textbf{Average} \\
\textbf{-1.8B} & 43.0 & 43.3 & 25.4 & 18.9 & 4.7 & 27.1 \\
\textbf{-7B} & 61.4 & 78.3 & 52.5 & 40.5 & 21.7 & 50.9 \\
\textbf{-20B} & 65.8 & 79.7 & 59.5 & 47.6 & 24.8 & 55.5 \\
\textbf{-Mixtral8x22B} & 77.5 & 82.0 & 63.6 & 50.3 & 36.8 & 62.0 \\
\bottomrule
\end{tabular}
\label{tab:plus_mathbench}
\end{table}

\end{document}